\documentclass[letterpaper, 10 pt, conference]{ieeeconf} 

\IEEEoverridecommandlockouts  

\usepackage{graphics} 
\usepackage{epsfig} 
\usepackage{mathptmx} 
\usepackage{times} 
\usepackage{amsmath} 
\usepackage{amssymb}  
\usepackage{multirow}
\usepackage{capt-of} 
\usepackage{makecell}
\usepackage{mathrsfs}
\DeclareMathAlphabet{\mathcal}{OMS}{cmsy}{m}{n}
\usepackage[caption=false,font=footnotesize]{subfig}
\usepackage[T1]{fontenc}
\usepackage{placeins}
\usepackage{cite}
\usepackage[caption=false,font=footnotesize]{subfig}

\title{\LARGE \bf
MinInter: Minimizing Trajectory Interpolation\\During Data Augmentation for Imitation Learning
}

\author{Qingyang~Wang, Xingang~Liu, Changwei~Yao, Zikai~Ouyang, Junwei~Liu, Haibo~Lu and Wei~Zhang%
\thanks{This work was supported in part by the National Science and Technology Major Project under Grant 2024ZD01NL00104; in part by the Major Key Project of PCL under Grant PCL2025A03; in part by the National Natural Science Foundation of China under Grant 62303208; and in part by the Science, Technology, and Innovation Commission of Shenzhen Municipality under Grant JCYJ20220530115010023.}%
\thanks{Equal contribution: Qingyang Wang, Xingang Liu, and Changwei Yao.}%
\thanks{Corresponding authors: Haibo Lu and Wei Zhang.}%
\thanks{Qingyang Wang, Zikai Ouyang and Wei Zhang are with the School of Automation and Intelligent Manufacturing, Southern University of Science and Technology, Shenzhen 518055, China, and also with Pengcheng Laboratory, Shenzhen 518000, China
{\tt\small \{wangqy2024, ouyang2022\}@mail.sustech.edu.cn, zhangw3@sustech.edu.cn}}%
\thanks{Xingang Liu and Junwei Liu are with the School of Automation and Intelligent Manufacturing, Southern University of Science and Technology, Shenzhen 518055, China
{\tt\small 12212110@mail.sustech.edu.cn, liujw@sustech.edu.cn}}%
\thanks{Changwei Yao is with Carnegie Mellon University, Pittsburgh, PA 15213, USA
{\tt\small changwey@andrew.cmu.edu}}%
\thanks{Haibo Lu is with Pengcheng Laboratory, Shenzhen 518000, China
{\tt\small luhb@pcl.ac.cn}}%
}

\begin{document}

\maketitle
\thispagestyle{empty}
\pagestyle{empty}


\begin{abstract}
Imitation learning enables robots to acquire complex manipulation skills from demonstrations, but its effectiveness is limited by the cost of collecting high-quality data. Trajectory-level data augmentation methods alleviate this challenge by recombining expert demonstrations under varied initial states. However, such methods typically insert interpolations or other non-expert transition segments between disjoint parts, and such non-expert segments could reduce the quality of the generated data. This paper introduces \textbf{Minimizing Interpolation (MinInter)}, an effective trajectory selection method that, for each sampled initial configuration, chooses the source demonstration requiring the least interpolation to form a complete trajectory. By explicitly minimizing interpolations during data generation, MinInter produces higher-quality synthetic demonstrations while remaining compatible with existing data generation frameworks. Experiments on 12 manipulation tasks with 26 variants from the MimicGen benchmark show that MinInter consistently improves both data generation success rates and policy success rates, with the largest gains on contact-rich, long-horizon and high-variance settings. Compared to the recent SkillGen framework, MinInter achieves higher policy success rates despite its conceptual simplicity, underscoring the value of interpolation minimization for data augmentation.
\end{abstract}


\section{INTRODUCTION}
Imitation learning has been widely applied in robotic manipulation, enabling robots to learn from pre-collected demonstrations. While imitation learning avoids the need for costly online interactions, its effectiveness critically depends on the availability of large-scale, high-quality demonstration data. However, collecting such data is time-consuming and expensive, limiting the performance and generalization of the learned policies. This makes data augmentation a key tool for expanding the demonstration dataset and 
improving policy robustness.

Recent work addresses data scarcity by trajectory-level augmentation, recombining expert demonstrations under randomized initial states to generate new trajectories~\cite{wang2024cyberdemo,mandlekar2023mimicgen,garrett2024skillmimicgen,jiang2024dexmimicgen}. These methods typically connect trajectory segments by introducing synthetic transitions (e.g., interpolations). However, such transitions are not demonstrated by experts and can introduce unnatural motions~\cite{garrett2024skillmimicgen,mandlekar2023mimicgen}, degrading the quality of the generated data and reducing downstream policy performance.

To address the above issue, we propose Minimizing Interpolation (MinInter), an effective trajectory selection method that reduces interpolations during data generation. For each sampled initial configuration, MinInter selects the source demonstration requiring minimal interpolation to form a complete trajectory, thereby enhancing the quality of synthetic data while remaining fully compatible with existing data generation frameworks.
We evaluate MinInter on all 26 variants of 12 open-source tasks from the MimicGen~\cite{mandlekar2023mimicgen} benchmark. Experimental results show that it consistently improves both policy success rates and data generation success rates, with particularly notable gains in contact-rich and long-horizon tasks. Compared with SkillGen~\cite{garrett2024skillmimicgen}, a recent state-of-the-art method built on the same baseline, MinInter achieves higher success rates while maintaining a streamlined design. These results demonstrate that reducing interpolation significantly enhances the quality of generated data and strengthens imitation learning performance.

Overall, we make the following contributions:
\begin{enumerate}
    \item We find that additional interpolation introduced during trajectory composition in data augmentation can degrade policy performance in imitation learning.
    \item We propose MinInter, a simple yet effective data augmentation method for imitation learning that selects trajectories with minimal total interpolation during data generation.
    \item We show that MinInter significantly improves the success rate of data
    generation, reducing the number of failed attempts during trajectory composition. 
    \item We evaluate MinInter on the MimicGen benchmark~\cite{mandlekar2023mimicgen}, where it consistently boosts policy success rates across nearly all tasks and further outperforms SkillGen~\cite{garrett2024skillmimicgen} on their reported subset, achieving state-of-the-art performance.
\end{enumerate}


\section{RELATED WORK}
\subsection{Data Augmentation for Imitation Learning}  
In imitation learning, data augmentation has been widely explored to improve policy performance by generating new training data from existing demonstrations~\cite{mitrano2022data,wang2024comprehensive}. Existing approaches can be broadly divided into image-level and trajectory-level augmentation.

Image-level augmentation operates on observations by applying transformations to image inputs. This includes basic operations such as rotation, color jitter, and cropping~\cite{zhan2021framework,laskin2020reinforcement,kostrikov2020image,young2021visual}, as well as more advanced generative or rendering-based approaches~\cite{ho2021retinagan,sinha2022s4rl,han2023expert,bharadhwaj2023roboagent,jin2025physically,zhang2024diffusion,black2023zero,mandi2022cacti,chen2023learning,chenrovi}. While these methods are effective for improving visual robustness, they do not directly address the physical structure or temporal organization of actions. Complementarily, this work focuses on trajectory-level augmentation, which modifies or generates trajectories to enhance the down stream policy learning.

Within trajectory-level augmentation, one line of work generates new trajectories using motion planning~\cite{dalal2023imitating,cheng2023nod,garrett2024skillmimicgen,wang2025hybridgen}, kinematic retargeting~\cite{yang2025physics}, generative models~\cite{jang2025dreamgen,yu2025real2render2real,wang2024robogen}, or counterfactual reasoning~\cite{pitis2020counterfactual,pitis2022mocoda,ameperosa2024rocoda}. These approaches broaden the available training data but often require additional modeling or high computational cost.
Another line of work creates new trajectories by recomposing segmented expert demonstrations. Frameworks such as MimicGen~\cite{mandlekar2023mimicgen} and DexMimicGen~\cite{jiang2024dexmimicgen} assemble subtask segments under randomized initial conditions and usually insert linear interpolation to connect misaligned parts. However, such transitions are not expert-demonstrated and may reduce trajectory quality. Our approach follows this setting but explicitly reduces interpolation during composition, thereby producing higher-quality data while remaining compatible with existing generation pipelines.

\vspace{-0.4mm}
\subsection{Transition Segments Handling}  
When generating new trajectories from existing expert demonstrations, the source demonstrations are first segmented and transformed to fit newly sampled initial states. After transformation, the segmented trajectories are often disconnected, requiring additional transition segments to connect adjacent transformed parts. Designing or handling these transition segments has therefore become an active research direction.

Several works propose different ways to form transition segments. Graph-based methods~\cite{yin2024graph} search for feasible paths through overlapping states; SkillMimic-V2~\cite{yu2025skillmimic} constructs a stitched trajectory graph to expand the coverage of the demonstration space; model-based approaches~\cite{hepburn2023model} synthesize connectors using learned dynamics models; and recovery-targeted methods~\cite{papagiannismiles,hoque2024intervengen} augment demonstrations by introducing corrective transitions to address distribution shift. While effective in connecting adjacent segments, these methods increase system complexity through model learning, planning, or manual interventions.

Other approaches explicitly employ interpolation as a form of transition segment. In this setting, end-effector poses are inserted between disconnected segments and a path is fitted through these poses to produce a continuous trajectory. 
CyberDemo~\cite{wang2024cyberdemo} introduces sensitivity analysis for trajectory segments to determine the extent of modification attributable to each action, thereby generating large-scale variation. Meanwhile, MimicGen~\cite{mandlekar2023mimicgen} and DexMimicGen~\cite{jiang2024dexmimicgen} apply linear interpolation to connect segments and have established a widely used benchmark. However, such interpolations are not demonstrated by experts and may degrade data quality. Existing methods do not directly address the negative effects of excessive transition segments on data quality and downstream policy learning. In contrast, we explicitly quantify the interpolation introduced during data generation and propose a method to minimize it, leading to improvements in both data generation and policy success rates.


\section{PROBLEM FORMULATION}
We consider the following data generation setup. 
Let $\mathcal{T}_{\text{exp}} = \{\tau_1, \dots, \tau_N\}$ denote a set of $N$ expert trajectories, where each trajectory $\tau_i$ can be segmented into $n$ object-centric segments, and each segment corresponds to a subtask involving interaction with a single static object unless the object is about to be grasped.
Segment $j$ of trajectory $\tau_i$ is represented as
\begin{equation}
\tau_i^{j} = \{\,E^{j}_{t,\text{src}}\,\}_{t=1}^{H_j},
\end{equation}
where $H_j$ is the number of timesteps in the segment, and $E^{j}_{t,\text{src}}$ denotes the end-effector pose at timestep $t$ in the world frame of the source trajectory.
The centric object associated with segment $j$ has pose $O^{j}_{\text{src}}$ in the source trajectory and $O^{j}_{\text{transf}}$ in the current subtask to be finished.

To generate new trajectories, we transform each segment by preserving the relative pose between the end-effector and the centric object.
Formally, for an initial state $s_0$, the transformed segment
$\tilde{\tau}_i^{j}(s_0) = \{\,E^{j}_{t,\text{transf}}\,\}_{t=1}^{H_j}$ is obtained by
\begin{equation}
(O^{j}_{\text{transf}})^{-1} E^{j}_{t,\text{transf}}
=
(O^{j}_{\text{src}})^{-1} E^{j}_{t,\text{src}}, 
\quad t=1,\dots,H_j .
\end{equation}
Rearranging gives the end-effector poses to be executed:
\begin{equation}
E^{j}_{t,\text{transf}}
=
O^{j}_{\text{transf}} \big(O^{j}_{\text{src}}\big)^{-1} E^{j}_{t,\text{src}} .
\end{equation}

A complete trajectory is obtained by concatenating the transformed segments
$\{\tilde{\tau}_i^{1}, \dots, \tilde{\tau}_i^{n}\}$.
Since adjacent segments are generally misaligned at their endpoints, transition segments are inserted between them using linear interpolation in both position and orientation. While necessary for continuity, such interpolations often introduce unnatural transitions.


\section{METHOD}
\label{sec:method}
    \setlength{\belowcaptionskip}{-10pt}

        \begin{figure*}
        \hfill
        \includegraphics[trim=0pt 10pt 0pt 10pt, clip, width=\linewidth]{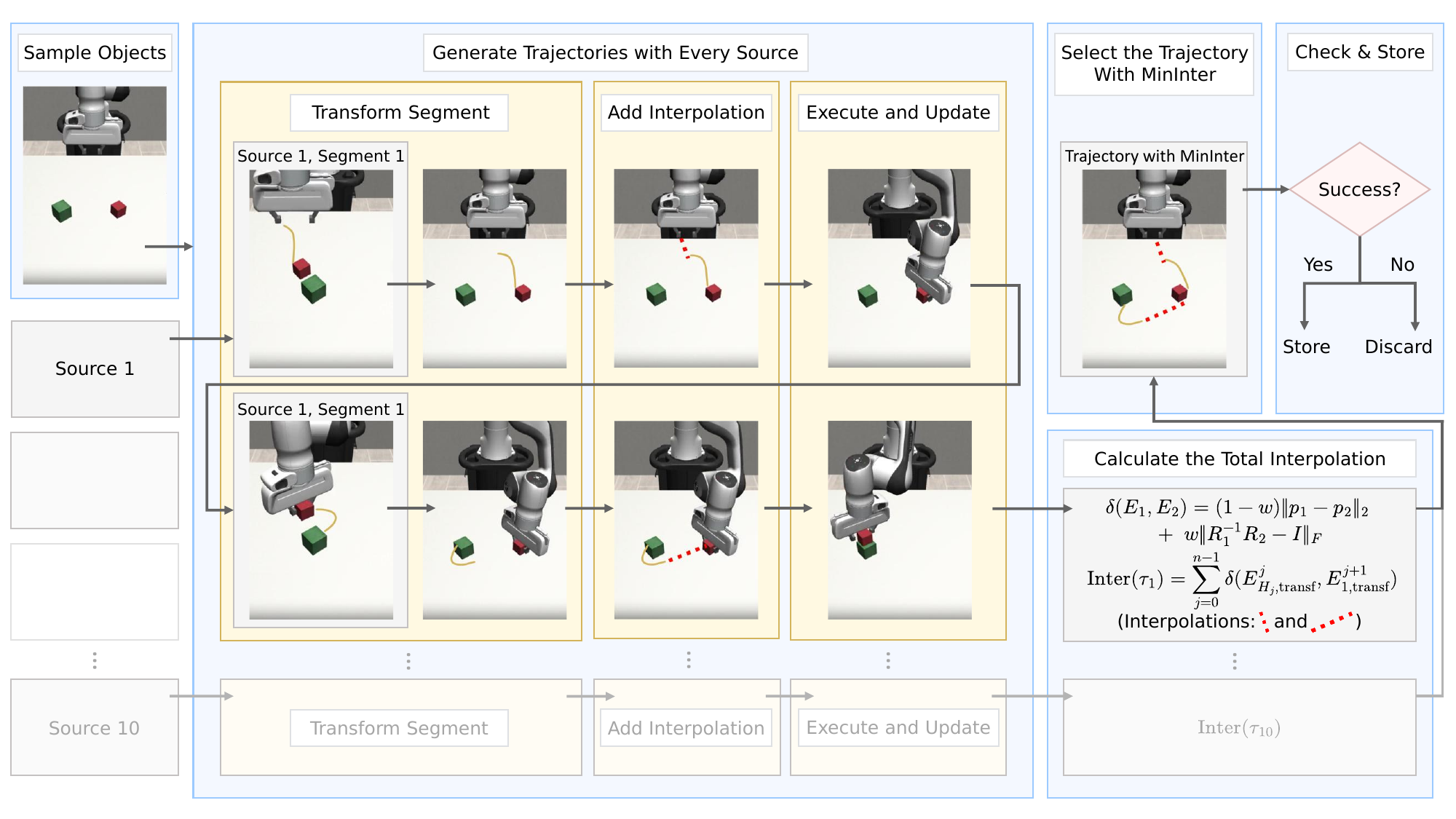}
        \caption{\textbf{Data generation pipeline with MinInter.} Illustration of the proposed MinInter method for selecting demonstration trajectories. For each sampled initial object configuration, trajectories are generated from all demonstrations by transforming their subtask segments and adding necessary interpolations. The total interpolation of each candidate is computed, and the trajectory with the smallest interpolation is selected and checked for task success. Only successful trajectories are added to the dataset.}
        \vspace{-2.0mm}
        \label{fig:algorithm}
    \end{figure*}
    
We propose Minimizing Interpolation (MinInter), a trajectory selection method that reduces interpolations introduced during trajectory composition. 
Given a sampled initial state, MinInter selects the source trajectory that yields the least total interpolation to form a complete trajectory.

\subsection{Interpolation Metric}  
Each end-effector pose $E \in \mathrm{SE}(3)$ consists of a position $p \in \mathbb{R}^3$ and a rotation $R \in \mathrm{SO}(3)$. 
To measure the interpolation between two poses $E_1$ and $E_2$, we introduce the metric
\begin{equation}
\delta(E_1,E_2) = (1-w)\|p_{1} - p_{2}\|_2 \;+\; w\|R_{1}^{-1}R_{2} - I\|_F , 
\end{equation}
where the weight $w \in [0,1]$ balances the contributions of translation and rotation, $\|\cdot\|_2$ denotes the Euclidean norm, and $\|\cdot\|_F$ the Frobenius norm. We empirically set $w = 0.5$, as it yields the best downstream policy performance in ablation experiments with varying weight values.
For a complete trajectory $\tau$ composed of $n$ segments, the total interpolation is calculated as the sum of interpolations between consecutive segments
\begin{equation}
\text{Inter}(\tau) = \sum_{j=0}^{n-1} \delta(E^{j}_{H_j,\text{transf}}, E^{j+1}_{1,\text{transf}}), 
\end{equation}
where $E^{j}_{H_j,\text{transf}}$ and $E^{j+1}_{1,\text{transf}}$ denote the end pose of segment $j$ and the start pose of the next segment $j+1$, respectively.
For $j=0$, $E^{0}_{H_0,\text{transf}}$ denotes the initial end-effector pose.

\subsection{MinInter Selection}  
Fig.~\ref{fig:algorithm} illustrates the overall data generation pipeline, with MinInter applied at the trajectory selection stage. Given a sampled initial state $s_0$, the objective is to generate a trajectory with minimal total interpolation by selecting an appropriate source trajectory. 
For consistency, all segments in a generated trajectory are taken from a single source trajectory, without combining segments from different sources. 
Formally, the selection problem is written as
\begin{equation}
\min_{\,\tau \in \mathcal{T}_{\text{exp}}} \; \text{Inter}(\Gamma(\tau, s_0)), 
\end{equation}
where $\Gamma(\tau, s_0)$ denotes the trajectory obtained by adapting $\tau$ to the initial state $s_0$. 
The optimal source trajectory is then
\begin{equation}
\tau^* = \arg\min_{\tau \in \mathcal{T}_{\text{exp}}} \; \text{Inter}(\Gamma(\tau, s_0)). 
\end{equation}
If $\tau^*$ satisfies the task-specific success condition, it is added to the training dataset; otherwise, a new initial state is sampled and the process repeats.


\section{Experiments}

\begin{figure*}[bt]
    \centering
    \setlength{\tabcolsep}{2pt}\renewcommand{\arraystretch}{0}
    \begin{tabular}{cccc}
        \subfloat[Stack]{\includegraphics[width=0.24\textwidth]{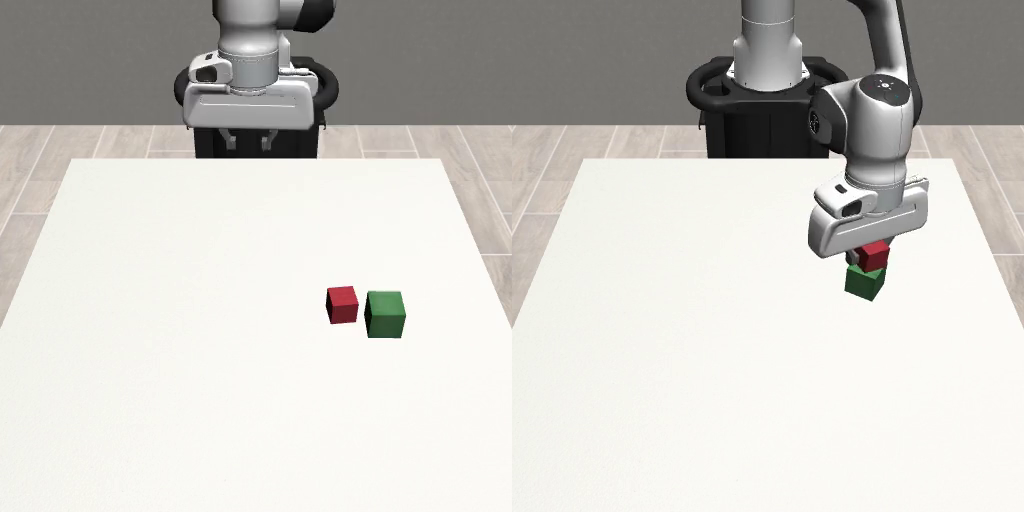}} &
        \subfloat[Stack Three]{\includegraphics[width=0.24\textwidth]{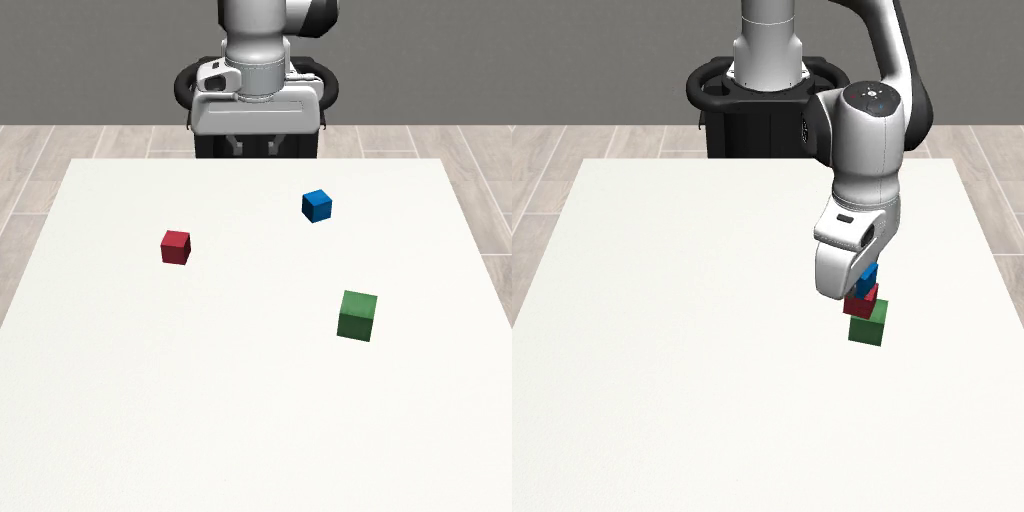}} &
        \subfloat[Square]{\includegraphics[width=0.24\textwidth]{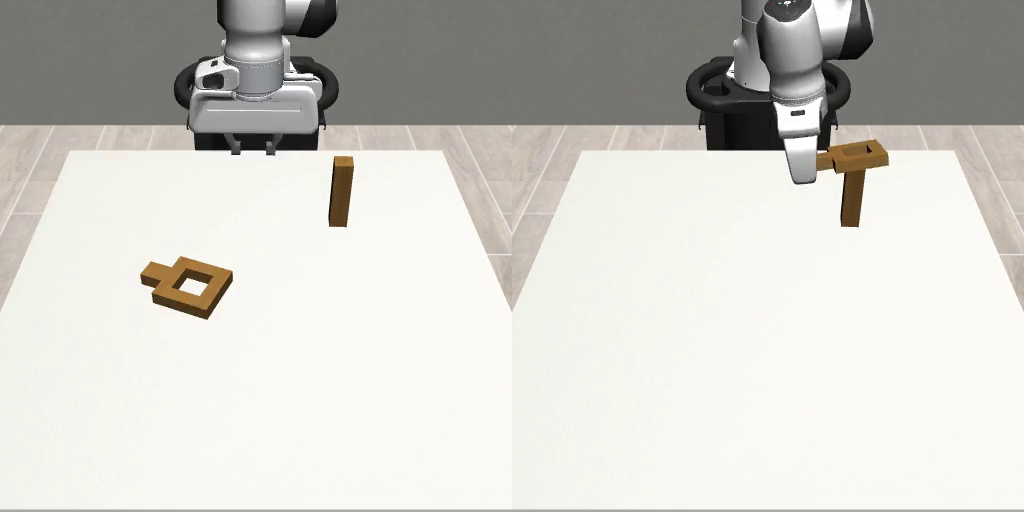}} &
        \subfloat[Threading]{\includegraphics[width=0.24\textwidth]{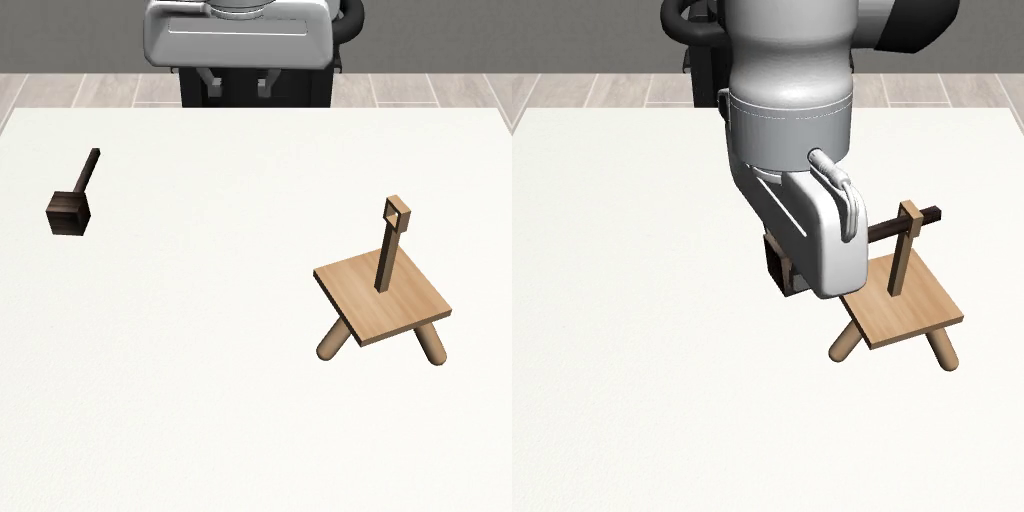}} \\
        \subfloat[Coffee]{\includegraphics[width=0.24\textwidth]{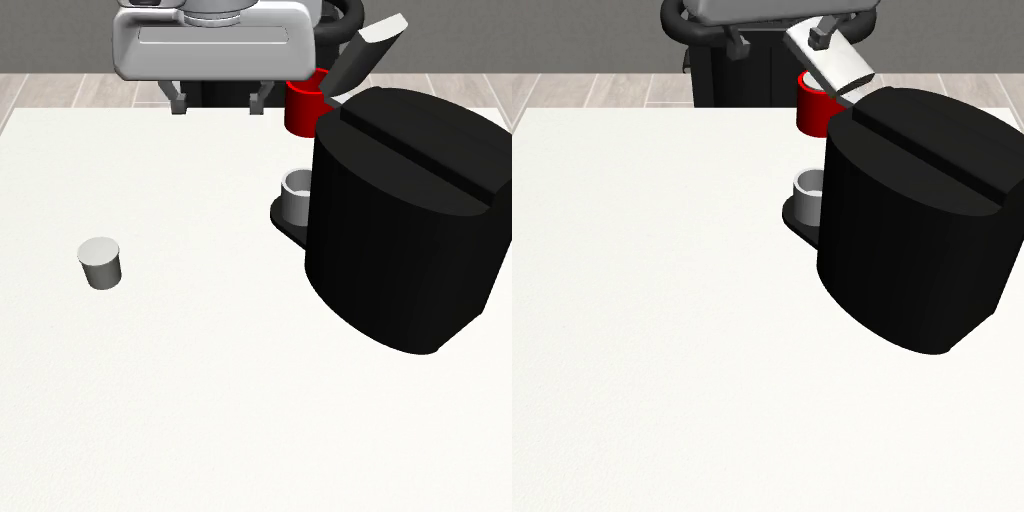}} &
        \subfloat[Three Pc.\ Assembly]{\includegraphics[width=0.24\textwidth]{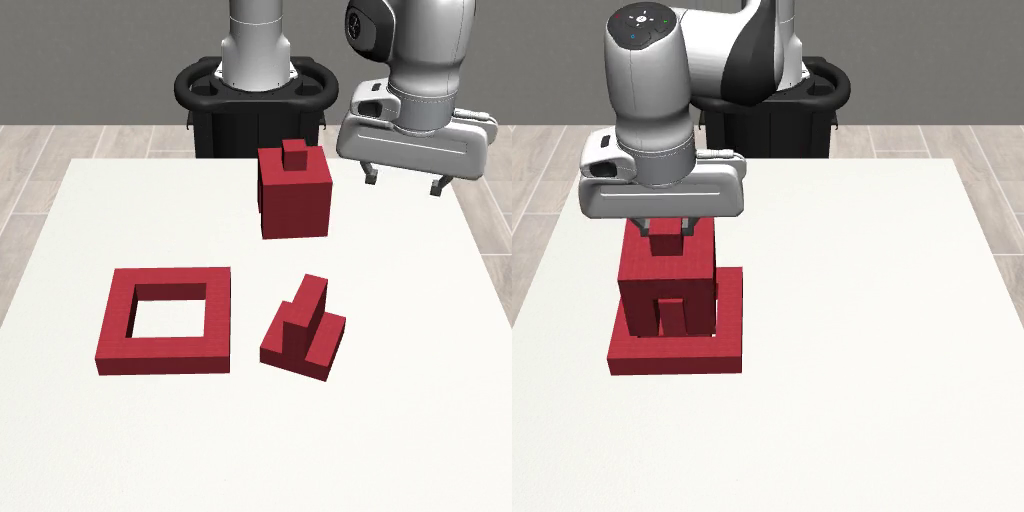}} &
        \subfloat[Hammer Cleanup]{\includegraphics[width=0.24\textwidth]{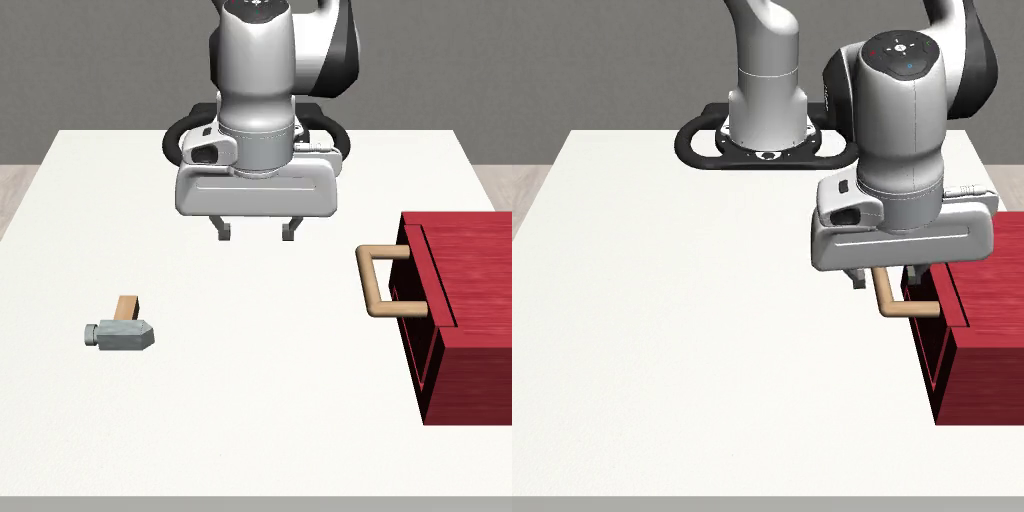}} &
        \subfloat[Mug Cleanup]{\includegraphics[width=0.24\textwidth]{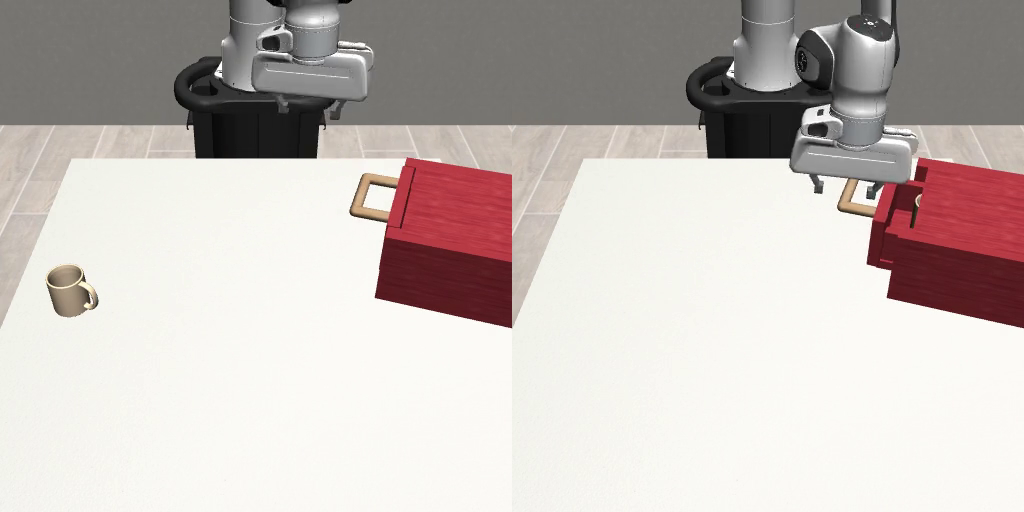}} \\
        \subfloat[Kitchen]{\includegraphics[width=0.24\textwidth]{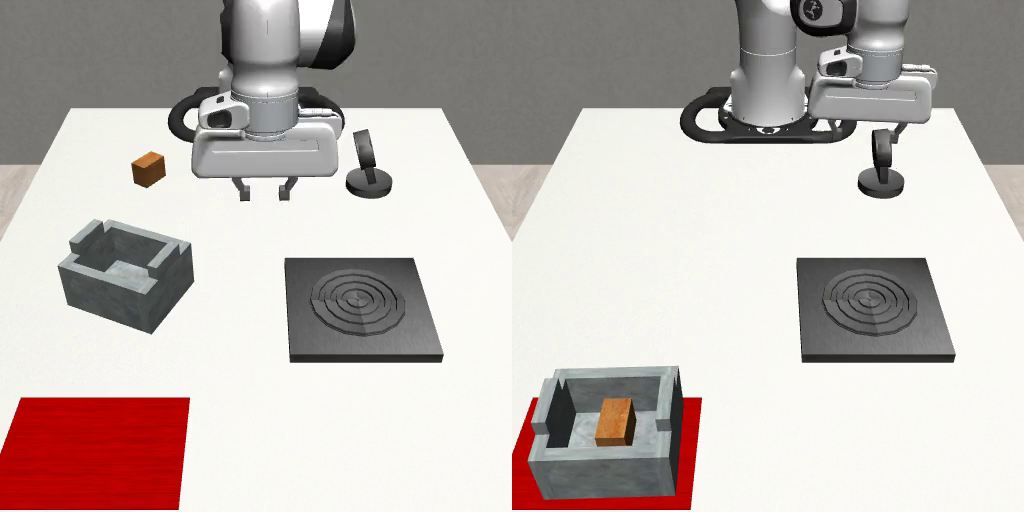}} &
        \subfloat[Nut Assembly]{\includegraphics[width=0.24\textwidth]{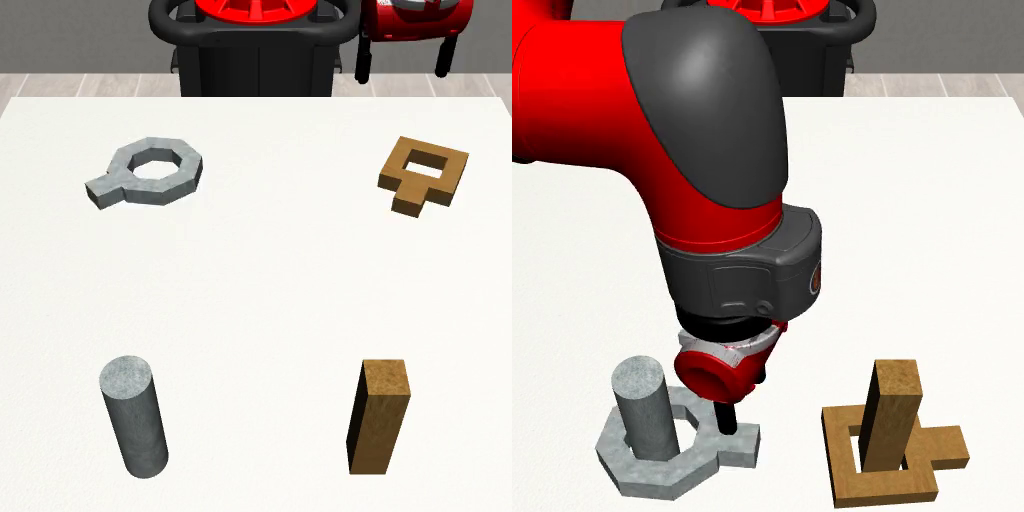}} &
        \subfloat[Pick Place]{\includegraphics[width=0.24\textwidth]{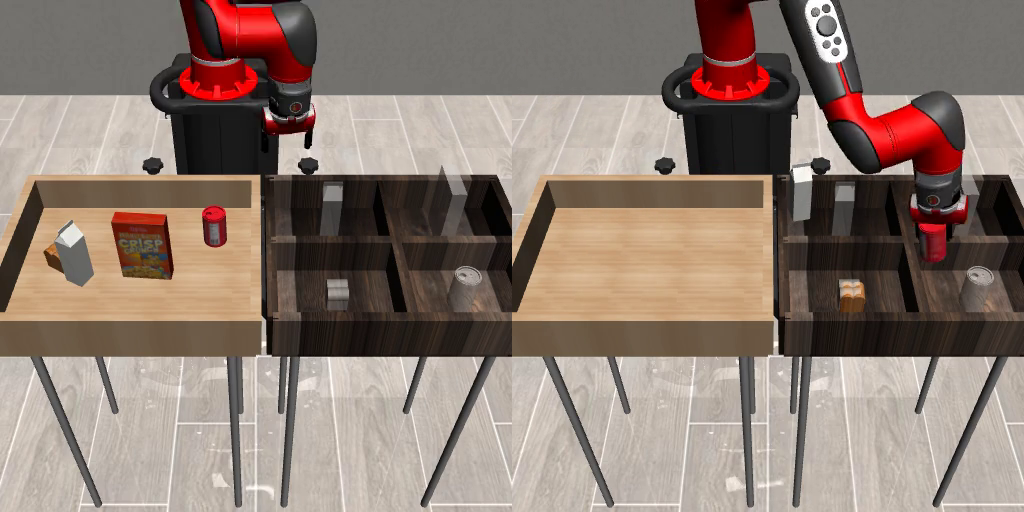}} &
        \subfloat[Coffee Preparation]{\includegraphics[width=0.24\textwidth]{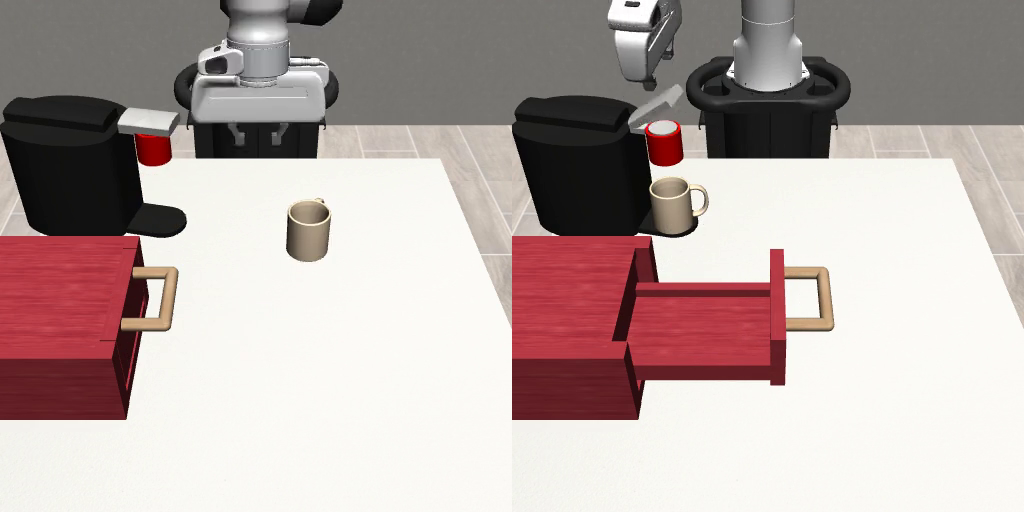}}
    \end{tabular}
    \caption{\textbf{Task visualization.} Visualization of the 12 manipulation tasks used in our experiments, sourced from the open benchmark~\cite{mandlekar2023mimicgen}. These tasks span a range of difficulty and behavior types, including basic tasks (a–b), contact-rich tasks (c–h), and long-horizon tasks (i–l). Environments are implemented using the MuJoCo simulator.}
    \label{fig:tasks}
    \vspace{-0.8mm}
\end{figure*}

We conduct experiments on a range of manipulation tasks using a unified benchmark to evaluate the effectiveness of MinInter. Specifically, we aim to answer two key questions:
Does MinInter consistently improve the success rates of downstream imitation learning policies?
And does it also enhance the success rate of data generation?

\subsection{Experimental Setup}
    \textbf{Tasks and Variants.} 
    We evaluate on all 12 publicly released MimicGen tasks with 26 variants (Fig.~\ref{fig:tasks}), spanning basic stacking (e.g., Stack, Stack Three), contact-rich alignment and articulation (e.g., Square, Threading), and long-horizon activities requiring sequential skills (e.g., Kitchen, Nut Assembly). To assess robustness under distribution shift, each task is run under three initial state distributions ($D_0$, $D_1$, $D_2$), with difficulty gradually increasing from $D_0$ to $D_2$ through more challenging object placements. 
    All experiments are conducted in robosuite~\cite{zhu2020robosuite} using MuJoCo.  
    
    \textbf{Data and Training.} 
    For each task, 10 expert demonstrations are used as source data to generate 1,000 successful trajectories per task variant, as determined by task-specific success criteria. These trajectories are used to train imitation learning policies with RGB observations (side-view and eye-in-hand) along with end-effector poses and gripper states. To ensure a fair comparison with the baseline~\cite{mandlekar2023mimicgen}, we adopt the same BC-RNN policy architecture and retain its evaluation protocol (50 rollouts every 20 epochs; best checkpoint per run), reporting the average over seeds 101, 102, and 103. Each policy is trained for 1200 epochs.

    \begin{figure}
        \centering
        \includegraphics[trim=0pt 10pt 0pt 10pt, clip, width=0.9\linewidth]{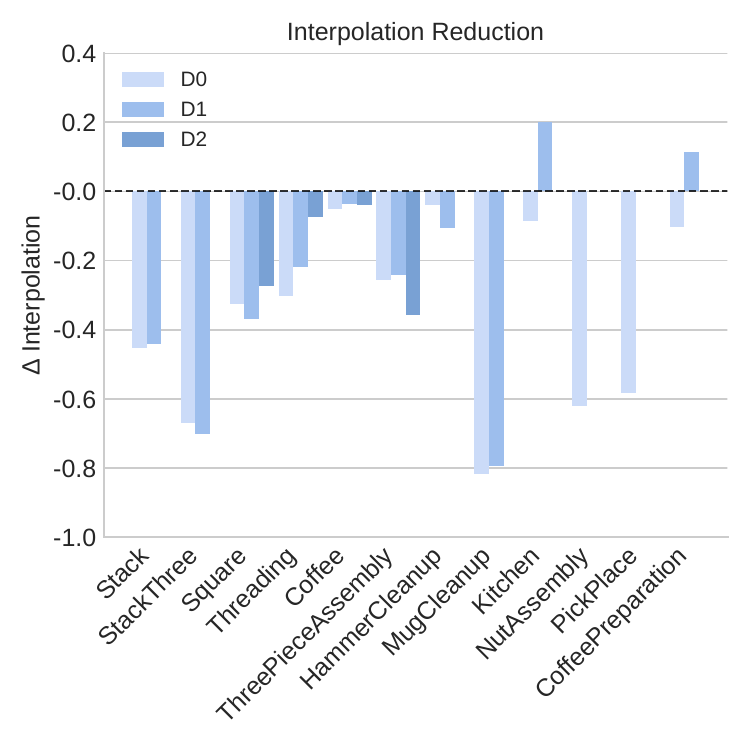}
        \caption{\textbf{Interpolation reduction.} Average change in interpolations between MinInter and the baseline across all evaluated task variants. Interpolations refer to the transition segments used in the baseline methods and are evaluated in translation and rotation as defined in our methodology. MinInter consistently reduces interpolations in the generated trajectories.}
        \vspace{-1.5mm}
        \label{fig:isd}
    \end{figure}

    \begin{figure}
        \centering
        \includegraphics[trim=0pt 10pt 0pt 10pt, clip, width=0.9\linewidth]{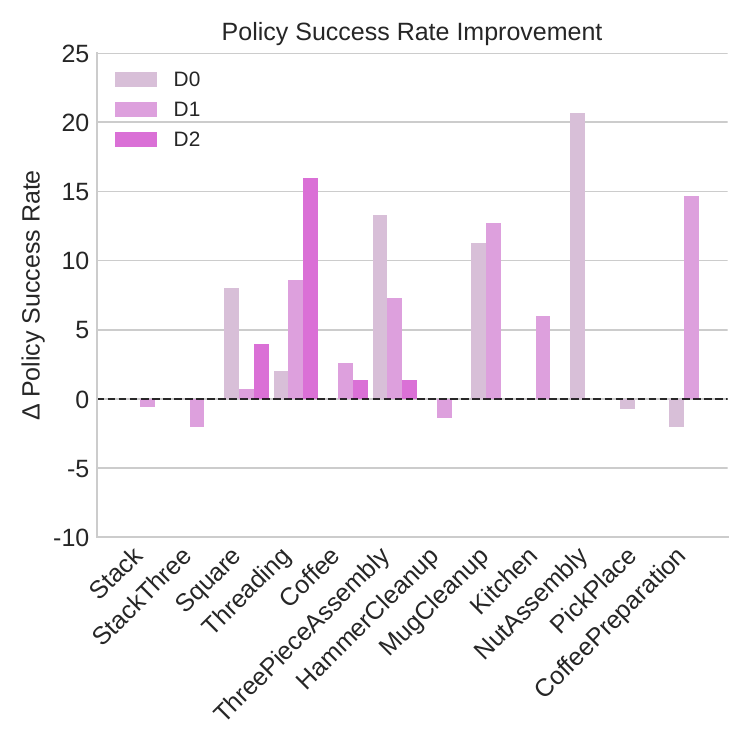}
        \caption{\textbf{Success rate improvement.} Change in policy success rate between MinInter and baseline across all evaluated task variants.}
        \vspace{-1.5mm}
        \label{fig:sr}
    \end{figure}

\subsection{Experimental Results}
    \textbf{Data Generation Success Rate.}
    According to Fig.~\ref{fig:isd}, MinInter produces trajectories with smaller interpolations than the baseline. Two exceptions occur in Kitchen $D_1$ and Coffee Preparation $D_1$, where slightly larger interpolations are observed; we attribute this to an increased data generation rate. Specifically, MinInter helps generate demonstrations in challenging configurations where the baseline fails due to large interpolation, thereby improving the success rate of the data generation process. 
    
    Table~\ref{tab:dgr_table} presents the detailed data generation success rates, showing that MinInter outperforms the baseline in most task variants, with an average gain of +12.8\% across tasks. Improvements are especially pronounced in contact-rich and long-horizon settings, such as Hammer Cleanup $D_0$ (+50.1\%) and Nut Assembly $D_0$ (+31.1\%), where generating valid trajectories is particularly challenging. 
    MinInter reduces failed attempts during data generation, thereby increasing the proportion of successful trajectories and enabling higher-quality demonstrations for training imitation learning policies.
    \begin{table*}[htb]
\begin{center}
\caption{\textbf{Policy Success Rate} }
\label{tab:sr_table}
\centering
\renewcommand{\arraystretch}{1.3}
\setlength{\tabcolsep}{5pt}
\begin{tabular}{|l|c|c|c|c|c|c|c|c|c|}
\hline
\multirow{2}{*}{\textbf{Task}} 
& \multicolumn{3}{c|}{\textbf{D0}} & \multicolumn{3}{c|}{\textbf{D1}} & \multicolumn{3}{c|}{\textbf{D2}} \\
\cline{2-10}
& \textbf{Base} & \textbf{SG} & \textbf{Ours} & 
\textbf{Base} & \textbf{SG}  & \textbf{Ours} &
\textbf{Base} & \textbf{SG} & \textbf{Ours} \\
\hline
Stack              & \textbf{100.0 $\pm$ 0.0} & - & \textbf{100.0 $\pm$ 0.0} & \textbf{99.3 $\pm$ 0.9} & - & 98.7 $\pm$ 0.9 & - & - & - \\
Stack Three        & \textbf{92.7 $\pm$ 1.9} & -  & \textbf{92.7 $\pm$ 3.8} & \textbf{86.7 $\pm$ 3.4} & - & 84.7 $\pm$ 1.9 & - & - & - \\
\hline
Square             & 90.7 $\pm$ 1.9& \textbf{100.0}  & 98.7 $\pm$ 0.9 & 73.3 $\pm$ 3.4& \textbf{84.0}  & 74.0 $\pm$ 2.8 & 49.3 $\pm$ 2.5 & \textbf{68.0} & 53.3 $\pm$ 3.8 \\
Threading          & 98.0 $\pm$ 1.6 & 94.0 & \textbf{100.0 $\pm$ 0.0} & 60.7 $\pm$ 2.5 & 46.0 & \textbf{69.3 $\pm$ 2.5} & 38.0 $\pm$ 3.3 & 34.0 & \textbf{54.0 $\pm$ 1.6} \\
Coffee             & \textbf{100.0 $\pm$ 0.0} & 98.0 & \textbf{100.0 $\pm$ 0.0} & 90.7 $\pm$ 2.5 & \textbf{100.0} & 93.3 $\pm$ 2.5 & 77.3 $\pm$ 0.9 & \textbf{92.0} & 78.7 $\pm$ 0.9 \\
Three Pc. Assembly & 82.0 $\pm$ 1.6 & 80.0 & \textbf{95.3 $\pm$ 1.9} & 62.7 $\pm$ 2.5 & 48.0 & \textbf{70.0 $\pm$ 1.6} & 13.3 $\pm$ 3.8 & \textbf{42.0} & 14.7 $\pm$ 1.9 \\
Hammer Cleanup     & \textbf{100.0 $\pm$ 0.0} & - & \textbf{100.0 $\pm$ 0.0} & \textbf{62.7 $\pm$ 4.7} & - & 61.3 $\pm$ 0.9 & - & - & - \\
Mug Cleanup        & 80.0 $\pm$ 4.9 & - & \textbf{91.3 $\pm$ 0.9} & 64.0 $\pm$ 3.3 & - & \textbf{76.7 $\pm$ 5.2} & - & - & - \\
\hline
Kitchen            & \textbf{100.0 $\pm$ 0.0} & - & \textbf{100.0 $\pm$ 0.0} & 76.0 $\pm$ 4.3 & - & \textbf{82.0 $\pm$ 1.6} & - & - & - \\
Nut Assembly       & 53.3 $\pm$ 1.9 & - & \textbf{74.0 $\pm$ 4.3} & - & - & - & - & - & - \\
Pick Place         & \textbf{50.7 $\pm$ 6.6} & - & 50.0 $\pm$ 4.9 & - & - & - & - & - & - \\
Coffee Preparation & \textbf{97.3 $\pm$ 0.9} & - & 95.3 $\pm$ 0.9 & 42.0 $\pm$ 0.0 & - & \textbf{56.7 $\pm$ 1.9} & - & - & - \\
\hline
\multicolumn{10}{|c|}{\textbf{MinInter vs. Baseline (12 tasks)}: \textbf{+4.8\%}} \\
\multicolumn{10}{|c|}{SkillGen vs. Baseline \textbf{(4 tasks)}: +3.8\%; MinInter on same 4 tasks: +4.9\%} \\
\hline
\end{tabular}
\end{center}
\footnotesize \textit{Note:} Baseline success rates are taken from its original paper, and SkillGen (SG) results from its appendix. Crucially, our experiments use the same source datasets and follow the same training and evaluation pipeline as the baseline and SkillGen, ensuring fully consistent experimental settings across all compared methods. 
A dash indicates that the task variant is not available in the open-source release. Compared to the baseline, MinInter consistently improves policy success rates with the largest gain reaching +20.7\%. It also outperforms SkillGen on their reported set, setting a new state of the art.
\vspace{-2.0mm}
\end{table*}
    \begin{table}[htb]
\caption{\textbf{Data Generation Success Rate} }
\label{tab:dgr_table}
\vspace{-20pt}
\begin{center}
\renewcommand{\arraystretch}{1.3} 
\setlength{\tabcolsep}{5pt}   
\begin{tabular}{|l|c|c|c|c|c|c|}
\hline
\multirow{2}{*}{\textbf{Task}} & 
\multicolumn{2}{c|}{\textbf{D0}} & 
\multicolumn{2}{c|}{\textbf{D1}} & 
\multicolumn{2}{c|}{\textbf{D2}} \\
\cline{2-7}
 & \textbf{Base} & \textbf{Ours} & 
   \textbf{Base} & \textbf{Ours} & 
   \textbf{Base} & \textbf{Ours} \\
\hline
Stack              & 0.943 & \textbf{0.953} & 0.900 & \textbf{0.903} & -     & -     \\
Stack Three        & \textbf{0.713} & 0.693 & \textbf{0.689} & 0.665 & -     & -     \\
\hline
Square             & 0.737 & \textbf{0.861} & 0.489 & \textbf{0.512} & 0.318 & \textbf{0.354} \\
Threading          & 0.510 & \textbf{0.652} & 0.392 & \textbf{0.446} & 0.216 & \textbf{0.300} \\
Coffee             & 0.782 & \textbf{0.816} & \textbf{0.635} & 0.628 & 0.277 & \textbf{0.338} \\
Three Pc. Assembly & 0.356 & \textbf{0.771} & 0.355 & \textbf{0.664} & 0.313 & \textbf{0.457} \\
Hammer Cleanup     & 0.476 & \textbf{0.977} & 0.204 & \textbf{0.386} & -     & -     \\
Mug Cleanup        & 0.295 & \textbf{0.450} & 0.170 & \textbf{0.307} & -     & -     \\
\hline
Kitchen            & \textbf{1.000} & \textbf{1.000} & 0.427 & \textbf{0.591} & -     & -     \\
Nut Assembly       & 0.500 & \textbf{0.811} & -     & -     & -     & -     \\
Pick Place         & 0.327 & \textbf{0.494} & -     & -     & -     & -     \\
Coffee Preparation & 0.532 & \textbf{0.625} & 0.361 & \textbf{0.597} & -     & -     \\
\hline
\multicolumn{7}{|c|}{\textbf{Overall average improvement: + 12.8\%}} \\
\hline
\end{tabular}
\end{center}
\footnotesize \textit{Note:} The data generation success rate is the proportion of generated trajectories satisfying the task-specific success criteria. MinInter achieves higher generation rates across most tasks, with particularly large gains in contact-rich and long-horizon settings.
\vspace{-1.9mm}
\end{table}

    \begin{figure}[t]
  \centering
  \subfloat[]{\includegraphics[width=0.45\linewidth]{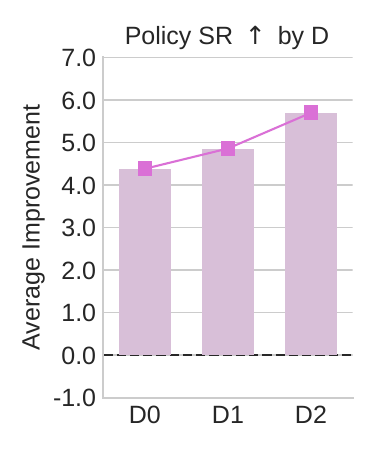}}\vspace{0.1em}
  \subfloat[]{\includegraphics[width=0.45\linewidth]{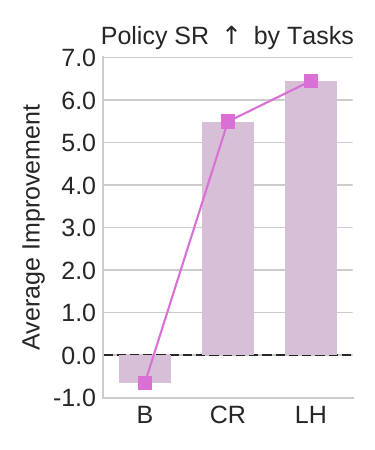}}
  \caption{\textbf{Policy Success rate improvement by task difficulty.} Average success rate improvement of MinInter over
 the baseline, grouped by two difficulty-related factors. (a) Reset distribution: D0 represents the default
 object initialization, D1 increases variation in position and orientation, and D2 introduces broader and more
 randomized sampling. (b) Task Group: Basic (B) tasks involve simple manipulation, Contact-Rich (CR) tasks
 require constrained physical interaction, and Long-Horizon (LH) tasks consist of multi-stage behavior sequences.
 The results show that MinInter achieves greater performance gains in more challenging settings.}
 \vspace{-2.5mm}
  \label{fig:sr_trend}
\end{figure}
    
    \textbf{Policy Success Rate.} 
    As shown in Fig.~\ref{fig:sr}, policies trained on MinInter-generated data outperform those trained on baseline data in the majority of cases, with full results reported in Table~\ref{tab:sr_table}. 
    Among the 26 task variants, 19 exhibit higher success rates or already reach 100\%, yielding an average improvement of +4.8\%. The largest gain is observed on Nut Assembly $D_0$, with an improvement of +20.7\%. 
    In contrast, some tasks already achieve strong baseline performance, leaving limited room for further improvement. For instance, Stack $D_0$ (100.0\%$\rightarrow$100.0\%) and Threading $D_0$ (98.0\%$\rightarrow$100.0\%) show only marginal gains, which consequently lowers the overall average.
    Furthermore, the improvements observed in policy success rates are consistent with the reduction in interpolations (Fig.~\ref{fig:isd}), suggesting that generating trajectories with fewer interpolations leads to more effective policies.

    \textbf{Comparison with State of the Art.} 
    We further compare our method with SkillGen~\cite{garrett2024skillmimicgen}, an influential approach for offline data augmentation. 
    Table~\ref{tab:sr_table} compares MinInter with SkillGen on the tasks evaluated in their paper
    with identical source data as well as training and evaluation codes adopted across all experiments.
    On this benchmark, SkillGen yields an average improvement of +3.8\% over the baseline, while MinInter achieves a higher improvement of +4.9\%, setting a new state of the art.
    Moreover, MinInter is easy to integrate, while remaining fully compatible with existing data generation frameworks.
    In contrast, SkillGen relies on fine-grained segmentation and requires separate skill and motion policies, adding additional implementation complexity.

    \textbf{Performance Across Tasks and Variants.} 
    We further analyze MinInter’s performance across different initial distributions and task groups, and find that its policy success rate improvements are more pronounced in challenging settings. As shown in Fig.~\ref{fig:sr_trend}, gains increase from $D_0$ to $D_2$, and are larger in contact-rich and long-horizon tasks than in basic tasks. These trends suggest MinInter is particularly effective when policies must learn from complex or diverse data.


\section{CONCLUSIONS}
We propose MinInter, a simple yet effective method that generates demonstrations with minimal interpolation, thereby improving data quality by reducing unnatural transitions. 
Experiments on 12 tasks with 26 variants show consistent gains in both data generation and policy success rates. 
The policy success rate improves across nearly all tasks and shows particularly strong gains on challenging tasks, achieving state-of-the-art performance on the evaluated benchmark. 
In future work, we plan to incorporate interpolation minimization into other data augmentation frameworks to further enhance data quality and advance imitation learning.


\section*{ACKNOWLEDGMENT}

We would like to thank the anonymous reviewers and our colleagues for their valuable comments, constructive suggestions, insightful discussions, and support, which helped improve this work.

\addtolength{\textheight}{-12cm}   

\FloatBarrier

\bibliographystyle{IEEEtran}
\bibliography{ref}

\end{document}